\documentclass[11pt]{article}

\usepackage[final]{acl}

\usepackage{times}
\usepackage{latexsym}
\usepackage{amsmath,amssymb,paralist}

\usepackage[T1]{fontenc}
\usepackage[utf8]{inputenc}

\usepackage{microtype}

\usepackage{inconsolata}

\usepackage{graphicx}
\usepackage{booktabs}
\usepackage{algorithm}
\usepackage{algorithmic}
\usepackage[dvipsnames]{xcolor}  % for \textcolor
\usepackage[breakable]{tcolorbox}
\usepackage{enumitem}  % for noitemsep, leftmargin
\usepackage{soul}
\usepackage{tikz}
\usepackage{float}
\usepackage{multirow}
\usepackage{subcaption}
\usetikzlibrary{arrows.meta,patterns}

\title{RADAR: \underline{R}etrieval-\underline{A}ugmented \underline{D}etector with \underline{A}dversarial \underline{R}efinement \\for Adaptive LLM-Generated Fake News Detection}

\author{
Song-Duo Ma \quad
Yi-Hung Liu \quad
Hsin-Yu Lin \\
\textbf{Pin-Yu Chen} \quad
\textbf{Hong-Yan Huang} \quad
\textbf{Shau-Yung Hsu} \quad
\textbf{Yun-Nung Chen} \\
National Taiwan University, Taipei, Taiwan \\
\texttt{\{r14944001, b11611049, r14922159\}@ntu.edu.tw}\\
\texttt{\{p14922004, r14922156, r14922090\}@ntu.edu.tw, y.v.chen@ieee.org}
}

\begin{document}
\maketitle

\begin{abstract}
To efficiently combat the spread of LLM-generated misinformation in the news domain, we present RADAR, a \textbf{R}etrieval-\textbf{A}ugmented \textbf{D}etector with \textbf{A}dversarial \textbf{R}efinement for adaptive LLM-generated fake news detection. Our approach employs a generator that rewrites real articles with factual perturbations, paired with a \emph{lightweight} detector that verifies claims using dense passage retrieval. To enable effective co-evolution, we introduce Verbal Adversarial Feedback (VAF). Rather than relying on scalar rewards, VAF issues structured natural-language critiques; these guide the generator toward more sophisticated evasion attempts, compelling the detector to adapt and improve. Experiments on an LLM-generated fake news benchmark show that RADAR outperforms retrieval-augmented trainable baselines and general-purpose LLMs with retrieval. Further analysis shows that retrieval on both the generator and detector sides improves performance, while VAF and few-shot demonstrations offer complementary benefits. RADAR also transfers better to fake news generated by an unseen external attacker, suggesting improved generalization beyond the specific co-evolved generator used during training.
\end{abstract}

\section{Introduction}

The proliferation of misinformation poses a significant threat to public discourse and democratic processes. Studies have shown that false news spreads farther and faster than accurate information on social media \cite{doi:10.1126/science.aap9559}, while large language models (LLMs) further lower the barrier to producing fluent, convincing fabricated content at scale \cite{zellers-etal-2019-defending}. Meanwhile, modern fake news detectors are typically fine-tuned encoder-based classifiers for efficiency and strong discrimination, yet they remain brittle under adversarial perturbations: small lexical edits or targeted factual modifications can sharply degrade performance \cite{jin2020bertreallyrobuststrong}. In practice, attackers iteratively probe detectors and refine rewrites, turning deployment into a moving target. This motivates continual adaptation: detectors should be updated to track emerging attack patterns while keeping the update process computationally efficient.

A natural approach is \emph{adversarial training}, where an attacker (generator) and a defender (detector) co-evolve. However, classic text GAN formulations such as SeqGAN \cite{yu2017seqgansequencegenerativeadversarial} rely on reinforcement learning to address non-differentiability, often leading to unstable training with high variance and mode collapse. More fundamentally, a detector typically returns only a scalar class confidence, indicating how likely an article is to be real, but offering little guidance on why a fake example was caught or how it should be revised. This limitation aligns with prior findings that language GANs often underperform simpler maximum-likelihood objectives \cite{caccia2020languagegansfallingshort}.

To make adversarial training practical for fake news detection, we avoid optimizing an end-to-end text-GAN objective. Instead, we use the generator as an adaptive attacker that produces factual perturbations, while training a lightweight detector that can be updated across adversarial rounds. This framing emphasizes two requirements: (i) a trainable detector that supports efficient round-wise adaptation, and (ii) generated fake articles that remain close to the source news in tone and event structure while altering specific factual claims.

In this work, we propose \textbf{RADAR}, a \textbf{R}etrieval-\textbf{A}ugmented \textbf{D}etector with \textbf{A}dversarial \textbf{R}efinement for adaptive LLM-generated fake news detection. RADAR studies a dual-retrieval adversarial training setup that grounds both the attacker and the defender in retrieved real-news evidence. On the generator side, retrieval provides journalistic style references that help preserve neutral news style in fact-altered rewrites. On the detector side, retrieval supplies external context for evidence-aware classification. The defender is implemented as a trainable encoder-based detector, which supports efficient round-wise updates and a lighter-weight alternative to prompt-only LLM judging.

To overcome the limitations of opaque scalar rewards, we introduce \textbf{Verbal Adversarial Feedback (VAF)}, which converts the detector’s judgment into structured natural-language diagnostics. Unlike a single probability score, VAF provides feedback including suspicious spans, detection rationales, and rewrite suggestions. Together with decision confidence and few-shot examples of successful attacks, these diagnostics give the generator explicit refinement cues for subsequent rewrites. The resulting refined examples are then used to further train the detector, supporting an iterative adversarial training loop.

Empirically, we evaluate RADAR on a fake news detection benchmark under controlled retrieval and feedback configurations. Our results show that RADAR generally improves across adversarial training rounds, suggesting that iterative refinement helps the detector adapt to emerging rewrite-style attacks. Ablation studies further show that retrieval, VAF, and few-shot demonstrations provide complementary benefits within this training loop. We also evaluate transfer to fake news generated by an unseen external attacker, providing initial evidence that the learned detector improves beyond the specific generator used during adversarial training.

Our contributions are 3-fold:
\begin{compactitem} 
\item We propose a dual-retrieval framework where generator-side retrieval supports realistic adversarial rewrites, and detector-side retrieval supports evidence-aware classification.
\item We develop a lightweight encoder-based detector that supports efficient round-wise adaptation and strong detection performance.
\item We introduce Verbal Adversarial Feedback (VAF), which converts detector decisions into structured critiques that identify suspicious spans, explain detection cues, and guide subsequent adversarial rewrites.
\end{compactitem}

\section{Related Work}

We review four lines of work related to RADAR: fake news detection, retrieval-augmented evidence for misinformation tasks, adversarial sample generation, and natural-language feedback for iterative refinement. These areas provide the basis for our study of retrieval and feedback in adaptive fake-news detection.

\paragraph{Fake News Detection}

Early fake news detection relied on linguistic cues and metadata such as source credibility.
\citet{wang2017liarliarpantsfire} introduced the LIAR benchmark and studied hybrid models combining text and metadata. With pre-trained language models, fine-tuned encoder-based classifiers became a dominant paradigm and substantially improved detection accuracy \cite{10.1007/s11042-020-10183-2}, with stronger backbones such as DeBERTa further boosting performance \cite{he2021debertadecodingenhancedbertdisentangled}. However, robustness remains challenging: small lexical edits or targeted modifications can significantly degrade classifier accuracy \cite{jin2020bertreallyrobuststrong}, and realistic threat models involve adaptive attackers that iteratively refine evasion attempts. Motivated by this gap, our work studies how a trainable detector can be updated under adaptive rewrite-style attacks.

\paragraph{Retrieval-Augmented Misinformation Detection}

Retrieval augmentation is widely used to support knowledge-intensive NLP by pairing parametric models with non-parametric memory. 
\citet{lewis2021retrievalaugmentedgenerationknowledgeintensivenlp} demonstrated that retrieval-augmented generation improves performance on tasks requiring external knowledge. This paradigm has also been adopted in misinformation detection and claim verification, where retrieval supplies supporting or refuting evidence to enable evidence-based decisions \cite{niu-etal-2024-veract}. In these settings, retrieved documents often serve as auxiliary context that improves verifiability and interpretability of predictions.

We further organize representative methods along two axes: whether retrieval is applied on the generator side as a realism prior, and whether retrieval is applied on the detector side as external evidence for verification. \autoref{fig:quadrant_related_work} summarizes this landscape and highlights RADAR as the only adversarial, dual-side design that uses generator-side retrieval for realism priors and detector-side retrieval for evidence-aware detection.

\begin{figure}[t]
\centering
\resizebox{\columnwidth}{!}{%
\begin{tikzpicture}[x=1cm,y=1cm]

% --- custom markers (clearer than bullets) ---
\newcommand{\advmark}{%
  \tikz[baseline=-0.6ex] \node[
    draw, fill=black, rectangle,
    minimum size=0.18cm, inner sep=0pt,
    sharp corners
  ] {};
}
\newcommand{\nonadvmark}{\tikz[baseline=-0.6ex]\node[draw,circle,minimum size=0.18cm,inner sep=0pt]{};}

% ---- canvas size ----
\def\xmin{-7.2}
\def\xmax{ 7.2}
\def\ymin{-5.6}
\def\ymax{ 5.9}

% ---- axes ----
% x-axis
\draw[very thick,-{Stealth[length=3mm]}] (\xmin+0.8,0) -- (\xmax-0.3,0);

% y-axis line (arrow head drawn LAST)
\draw[very thick] (0,\ymin+0.5) -- (0,\ymax-0.55);

% ---- +RAG / -RAG labels ----
\node[font=\Large\bfseries, fill=white, inner sep=2pt] at ( \xmax-1.5, 0.55) {+RAG (G)};
\node[font=\Large\bfseries, fill=white, inner sep=2pt] at ( \xmin+2, 0.55) {-RAG (G)};

\node[font=\Large\bfseries, fill=white, inner sep=2pt] at ( 1.2, \ymax-1) {+RAG (D)};
\node[font=\Large\bfseries, fill=white, inner sep=2pt] at ( 1.2, \ymin+1) {-RAG (D)};

% ---- axis titles ----
\node[font=\LARGE\bfseries, align=center, fill=white, inner sep=3pt]
  at (0,\ymin-0.25) {Retrieval Augmentation\\on the Generator};

\node[font=\LARGE\bfseries, rotate=90, align=center]
  at (\xmin+0,0) {Retrieval Augmentation\\on the Detector};

% ---- legend ----
\node[anchor=north east, align=left, font=\normalsize,
      fill=white, draw=black!30, inner sep=3pt] at (\xmax,\ymax-0.3) {%
\begin{tabular}{@{}l@{}}
\advmark~Adversarial\\
\nonadvmark~Non-adversarial
\end{tabular}
};

% ---- quadrant items (cite only) ----
\node[anchor=west, align=left, font=\Large] at (-6.2, 3) {%
\advmark~\citet{chen-etal-2025-real}\\[2pt]
\nonadvmark~\citet{niu-etal-2024-veract}\\[2pt]
\nonadvmark~\citet{nezafat-samet-2024-rag-genai-fake-news}%
};

\node[anchor=west, align=left, font=\Large\bfseries,
      fill=black!12, rounded corners=2pt, inner sep=3pt]
  at (0.7, 3.7) {%
\advmark~RADAR (Ours)%
};

\node[anchor=west, align=left, font=\Large] at (-6.2,-2.1) {%
\advmark~\citet{zellers-etal-2019-defending}\\[2pt]
\advmark~\citet{tian-etal-2025-symbolic}\\[2pt]
\advmark~\citet{wang-etal-2025-llm-gan}%
};

\node[anchor=west, align=left, font=\Large] at (0.7,-1.45) {%
\advmark~\citet{singh-namin-2024-adversarial-rag-fake-news}%
};

% y-axis arrow head (draw LAST so it stays visible)
\draw[very thick,-{Stealth[length=3mm]}] (0,\ymax-0.55) -- (0,\ymax-0.4);

\end{tikzpicture}%
}
\caption{The x-axis indicates generator-side retrieval, and the y-axis indicates detector-side retrieval. Markers distinguish adversarial methods from non-adversarial ones. Our RADAR is an adversarial, dual-side design.}
\label{fig:quadrant_related_work}
\end{figure}

\paragraph{Adversarial Sample Generation and Text GANs}

Text GANs adapt the GAN paradigm \cite{goodfellow2014generativeadversarialnetworks} to discrete language, but optimization is difficult due to non-differentiable token sampling. SeqGAN \cite{yu2017seqgansequencegenerativeadversarial} addresses this via policy-gradient reinforcement learning, and later variants such as LeakGAN \cite{guo2017longtextgenerationadversarial} and RankGAN \cite{lin2018adversarialrankinglanguagegeneration} explore richer training signals and objectives. However, RL-based text GANs often suffer from high variance and unstable training, and can be outperformed by maximum-likelihood trained language models in quality--diversity trade-offs \cite{caccia2020languagegansfallingshort}. Rather than optimizing an end-to-end text-GAN objective, we use the generator as an adaptive attacker and replace scalar rewards and policy gradients with structured natural-language feedback from the detector.

\paragraph{Natural Language Feedback for Optimization}

A growing body of work uses natural language as an optimization signal. Self-Refine \cite{madaan2023selfrefineiterativerefinementselffeedback} shows that a model can critique and iteratively improve its own output without parameter updates. Reflexion \cite{shinn2023reflexionlanguageagentsverbal} accumulates verbal reflections across agentic episodes as a form of semantic memory. Constitutional AI \cite{bai2022constitutionalaiharmlessnessai} steers generation toward safety via rule-based critiques. More recently, TextGrad \cite{yuksekgonul2024textgradautomaticdifferentiationtext} formalizes natural language feedback as ``textual gradients,'' showing that LLM-generated critiques can propagate through complex systems analogously to backpropagation.

These methods typically use feedback in a cooperative setting, where critiques help a model improve toward a fixed target. RADAR instead uses feedback in an adversarial training loop: detector-side signals are exposed to the generator to guide subsequent rewrites, and the resulting examples are then used to update the detector. This distinction changes the role of feedback from self-improvement toward generating training examples that reflect failure modes observed by the detector.

\section{Methodology}

Given a news article $x$, the detector predicts a binary label $y \in \{0,1\}$ indicating whether $x$ is real or fake. We focus on adaptive rewrite-style attacks, where a fake article preserves the event frame and journalistic tone of a real source article while altering specific factual claims. RADAR addresses this setting through a round-wise adversarial training framework between a generator $G$ and a detector $D$, with retrieval optionally enabled on either side.

\begin{figure*}[t]
\centering
\includegraphics[width=\textwidth]{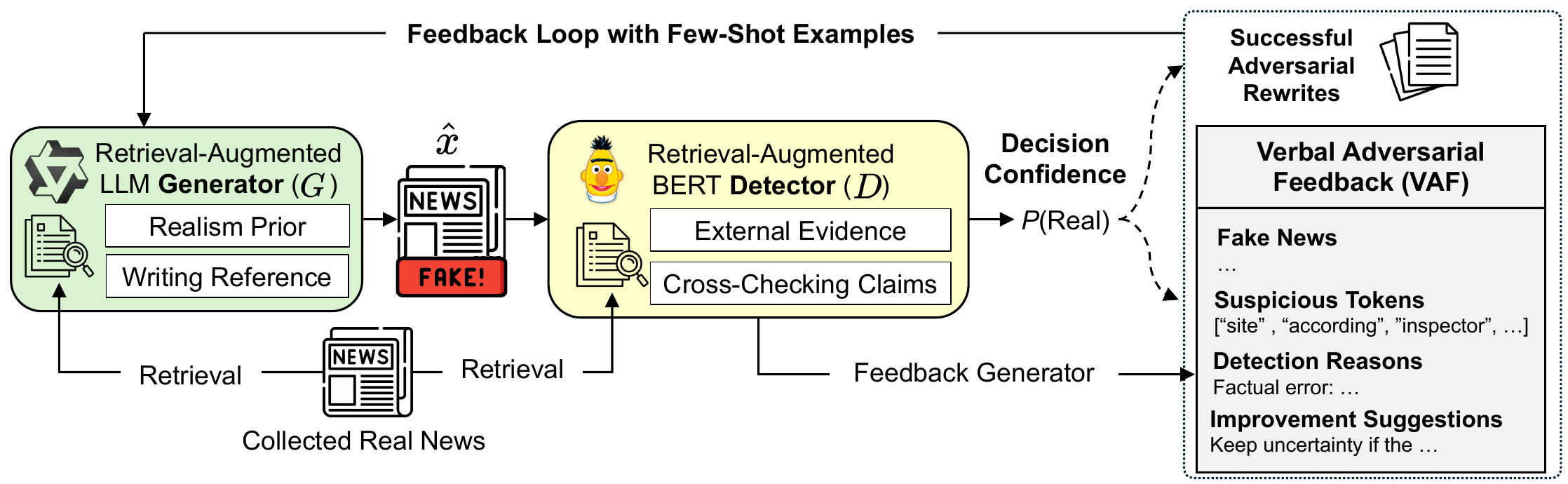}
\caption{Overview of RADAR for LLM-generated fake news detection under adaptive attacks. Retrieval provides a realism prior for the generator and external evidence for the detector. The detector provides confidence scores and diagnostic signals for VAF, while successful attacks are cached as few-shot examples to guide generator refinement.}
\label{fig:intro}
\end{figure*}

\subsection{RADAR Framework}

\autoref{fig:intro} illustrates the RADAR pipeline. At round $t$, the generator rewrites a real news article $x_i$ into a fact-altered variant $\hat{x}_{i,t}$, and the detector evaluates whether the article is real or fake. Detector-side diagnostics are then converted into verbal adversarial feedback (VAF), which is stored and used to condition the generator in the next round.

Retrieval can be enabled independently for the generator and the detector, yielding four configurations: $G^{-}/D^{-}$, $G^{+}/D^{-}$, $G^{-}/D^{+}$, and $G^{+}/D^{+}$. This decomposition allows us to separately study the effect of generator-side retrieval, detector-side retrieval, and their combination.

Across rounds, RADAR updates the detector using newly generated real–fake pairs, while the generator is refined through prompt-time feedback, cached successful rewrites, and periodic parameter-efficient updates. This round-wise design allows us to examine whether adversarial refinement improves detector robustness over successive rounds.

\subsubsection{Generator}

The generator $G$ is an instruction-tuned LLM that rewrites a real news article into a deceptive fake variant designed to evade detection. Given a real news article $x$, retrieval context $c$, verbal adversarial feedback from the previous round $\mathrm{VAF}_{t-1}$, and cached few-shot exemplars $\mathcal{E}_{t-1}$, the generator produces $\hat{x}=G(x,c,\mathrm{VAF}_{t-1},\mathcal{E}_{t-1})$.

When generator-side retrieval is enabled, the retrieved passages are used only as realism priors and writing references, rather than as facts to preserve. They provide cues about journalistic style, event framing, and plausible levels of detail, helping the generator produce more natural rewrites.

The generator is instructed to modify the article’s underlying facts while preserving a realistic news style. To induce diverse attack patterns, each generation is conditioned on a manipulation strategy randomly sampled from five options: (i) entity substitution; (ii) numerical or temporal distortion; (iii) outcome reversal; (iv) false causal attribution; or (v) fabricated supporting detail injection.

\subsubsection{Detector}

The detector $D$ is a trainable encoder-based classifier that predicts whether an input article is real or fake. Given an article $x$ and retrieval context $c$, the detector outputs a decision confidence $P(\mathrm{real}\mid x,c)$ indicating the likelihood that the article is real. When detector-side retrieval is enabled, the retrieved passages are used as external evidence for cross-checking the claims in the input article.

A scalar decision confidence alone provides limited information about why a generated article was detected or how $G$ should revise it. We therefore augment the detector with VAF, described below.

\subsubsection{Verbal Adversarial Feedback}

Verbal Adversarial Feedback (VAF) is a structured diagnostic signal composed of three components: \textit{suspicious tokens}, \textit{detection reasons}, and \textit{improvement suggestions}. Instead of using the detector output only as a scalar confidence score, VAF exposes interpretable detector-side diagnostics that can be reused to guide subsequent adversarial rewriting.

VAF is constructed in two steps. First, we extract Gradient-Identified Suspicious Tokens from the detector to identify input spans that strongly influence the prediction. Second, a dedicated Feedback Generator verbalizes these detector-side signals into detection reasons and actionable rewrite suggestions, as illustrated in \autoref{tab:vaf}.

\begin{table}[t]
\centering
\fbox{%
  \begin{minipage}{0.95\linewidth}
  \footnotesize
  \raggedright
  \setlength{\baselineskip}{13pt}
  \textbf{Original Real News}\\
  ``The cause of the blast was not immediately known, a policeman at the site said, adding that a bomb squad was on its way to investigate.''\\[3pt]

  \textbf{Generated Fake News}\\
  ``The blast was caused by a faulty gas line that had been improperly sealed during a recent renovation, according to a city inspector on-site.''\\[-1pt]

  \noindent\rule{\linewidth}{0.4pt}\\[-1pt]
  
  \textbf{Suspicious Tokens}\\
  \{\texttt{site}, \texttt{according}, \texttt{inspector}, \texttt{blast}, \texttt{renovation}\}\\[3pt]

  \textbf{Detection Reasons}\\
  Factual error: the generated article invents a known cause, while the original states that the cause was unknown.\\
  Contextual cue: the fake article relies on vague attribution to a city inspector and introduces unsupported causal details.\\[3pt]

  \textbf{Improvement Suggestions}\\
  Keep uncertainty if the investigation is ongoing; cite a named authority only if the causal claim is verifiable.
  \end{minipage}%
}
\caption{An example of VAF produced for a generated fake news article. The feedback identifies both the factual perturbation and the contextual cues that make the rewrite detectable.}
\label{tab:vaf}
\end{table}

\paragraph{Gradient-Identified Suspicious Tokens}
We extract suspicious tokens with integrated gradients (IG), which assigns token-level attribution scores by integrating gradients of the detector output with respect to the input embeddings from a baseline to the observed input. After filtering stopwords, we retain the top five tokens with the highest attribution scores as suspicious tokens. These tokens serve as lexical cues for subsequent feedback generation.

\paragraph{Feedback Generator}
We use an LLM-based generator to instantiate the detection reasons and improvement suggestions in VAF. Its input includes the source article, the generated fake article, the detection confidence, and the suspicious tokens with their attribution scores. From these signals, it produces structured natural-language feedback that summarizes likely detection errors and proposes targeted revisions for subsequent rewriting. This feedback is then injected into the generator prompt in the next round to support context-aware adversarial refinement.

\subsection{Training Strategy}
\label{sec:training}

RADAR is trained in rounds with alternating updates of the detector and generator. Each round includes adversarial generation, detector training on real--fake pairs, and generator adaptation using successful attacks. The complete procedure is given in Algorithm~\ref{alg:training} in \autoref{sec:appendix}.

\paragraph{Detector Training}
The detector is trained as a supervised binary classifier over real and generated fake articles. In each adversarial round, we continue training the detector from the previous round’s checkpoint on a balanced set of real articles and newly generated fake articles, labeled as $y=1$ and $y=0$, respectively. For an input article $x_i$ with detector-side retrieval context $c_i$, the detector outputs $p_i=P(\mathrm{real}\mid x_i,c_i)$ and is optimized with binary cross-entropy:
\begin{equation}
\mathcal{L}_D = -\sum_i \left[y_i \log p_i + (1-y_i)\log (1-p_i)\right],
\end{equation}
where $y_i \in \{0,1\}$ is the ground-truth label.

Cached successful attacks are used only on the generator side, either as few-shot exemplars for later prompts or as supervised targets for LoRA adaptation. The detector remains a supervised encoder classifier trained on labeled real--fake pairs throughout all rounds.

\paragraph{Generator Training}
We refine the generator without policy-gradient optimization. Instead, RADAR adapts the generator through three mechanisms: (i) VAF-conditioned prompting for immediate round-to-round refinement, (ii) cached successful adversarial rewrites as few-shot exemplars, and (iii) periodic LoRA-based supervised fine-tuning for persistent adaptation.

At each round, we distinguish between the threshold used to define attack success and the stricter threshold used for generator adaptation. A generated fake article is considered a successful attack if it fools the detector, i.e., $P(\mathrm{real}\mid \hat{x})>0.5$. For few-shot caching and LoRA fine-tuning, we use a stricter threshold of $0.6$ to retain only high-confidence successful attacks and avoid adapting the generator to borderline examples near the detector's decision boundary. We set this threshold a priori as a conservative margin above the binary decision threshold of $0.5$. At each round, we retain the top-$m$ high-confidence examples ranked by detector confidence as an adaptation pool, from which one example is randomly sampled for a single LoRA update.

The generator is optimized with
\begin{equation}
\mathcal{L}_G = \mathcal{L}_{\mathrm{CE}} + \lambda_{\mathrm{KL}} D_{\mathrm{KL}}(\pi_\theta \| \pi_{\mathrm{ref}}),
\end{equation}
where $\mathcal{L}_{\mathrm{CE}}$ is the next-token cross-entropy loss on the sampled successful rewrite, $\pi_\theta$ is the current LoRA-adapted generator, and $\pi_{\mathrm{ref}}$ is the frozen base generator before LoRA adaptation. The KL term regularizes the adapted generator toward the base model to limit distribution drift during repeated adversarial updates.

\section{Experiments}

In this section, we evaluate RADAR and address four questions: (1) whether RADAR improves fake-news detection over zero-shot LLM detectors and trained baselines, (2) how generator-side and detector-side retrieval contribute to performance, (3) how the proposed feedback design affects robustness, and (4) whether the learned detector generalizes to attacks from an unseen external attacker. We provide additional analyses in \autoref{app:additional_analysis}, including retrieval-overlap analysis and the effect of retrieval strategy and depth.

\subsection{Experimental Setup}

\paragraph{Dataset and Retrieval Index}
We conducted our experiments using the \textit{AdvFake-News-Please} dataset released by \citet{chen-etal-2025-real}, which contains real news articles sourced from the NewsPlease repository along with their corresponding adversarial fakes. To support retrieval augmentation, we use the same public news database as the retrieval corpus, consisting of approximately 811K news articles collected from multiple news sources within the same date range. Following \citet{chen-etal-2025-real}, we filter the retrieval corpus to exclude exact duplicates of the seed news articles and remove articles from the source outlet of the seed articles, reducing the risk of trivial retrieval leakage.

We build a dense FAISS~\cite{douze2024faiss} index using off-the-shelf DPR checkpoints (\texttt{dpr-question\_encoder-single-nq-base} and
\texttt{dpr-ctx\_encoder-single-nq-base} for query and passage encoding, respectively) ~\cite{karpukhin2020densepassageretrievalopendomain}.
During adversarial rounds, generator-side retrieval uses the source real article $x_i$ as the query. Detector-side retrieval instead uses the article currently being classified: $\hat{x}_{i,t}$ for generated fake articles and $x_i$ for real articles. In both cases, we retrieve the top-3 most semantically similar passages, following our retrieval-depth analysis in \autoref{app:retrieval_strategy_depth}, where DPR with $k=3$ consistently outperforms $k=1$ and $k=5$ across all three generator backbones.

\paragraph{Training Protocol}
Following Algorithm~\ref{alg:training}, the generator uses a fixed set of 500 real seed articles across all 5 rounds, producing 500 fake articles per round. The detector is trained each round on a balanced set of 500 newly sampled real articles and 500 generated fakes, resulting in 1,000 instances per round and 5,000 in total. We adopt the dataset’s standard split and fixed hyperparameters throughout, without additional round-wise validation tuning, to isolate the effect of co-evolution.

\paragraph{Generator Implementation}
We instantiate the generator $G$ with three compact instruction-tuned LLM backbones: \textbf{Qwen3-4B-Instruct} \cite{yang2025qwen3}, \textbf{Gemma3-4B} \cite{gteam2025gemma3}, and \textbf{Phi-3.5-mini (3.8B)} \cite{abdin2024phi3}. Each generator is fine-tuned separately under the same adversarial training pipeline. For parameter-efficient and stable adaptation, we use LoRA with rank $r=16$, scaling factor $\alpha=32$, and dropout $0.05$. A generated fake article is considered a successful attack if $P(\text{real}) > 0.5$, while a stricter threshold of $P(\text{real}) > 0.6$ is used for generator adaptation and exemplar caching. At each round, we rank these high-confidence attacks by detector confidence and retain the top 3 as the adaptation pool. One example is then randomly sampled from this pool for LoRA adaptation. The retained attacks are also added to an exemplar cache of size 5, which keeps only the most recent examples; one cached example is used as a few-shot demonstration in subsequent generation. LoRA adaptation is performed after every round using a single optimizer step with batch size 1. We optimize the generator with AdamW using a learning rate of $1 \times 10^{-4}$ and apply a KL-divergence penalty ($\lambda=0.01$) to limit drift from the base model distribution.

For feedback generation, we reuse the same underlying LLM instance but disable the LoRA adapters during inference, so that feedback is produced using the frozen base-model weights. A separate feedback-generation prompt is used, and no parameters are updated during feedback generation. We set the feedback-generation threshold to $\tau_{\mathrm{fb}}=0.5$: detailed VAF is generated when $P(\text{real}) \leq 0.5$, while samples above this threshold receive score-only feedback.

\paragraph{Detector Implementation}
The detector $D$ is based on \textbf{DeBERTa-v3-base}~\cite{he2021debertadecodingenhancedbertdisentangled}, fine-tuned for binary classification (Real vs.\ Fake). We use this backbone because it is lightweight and efficiently updatable for evolving attacks, while also reducing the risk of pretraining-time contamination from 2024-era evaluation articles relative to more recent encoders. The detector is trained with AdamW using a learning rate of $5 \times 10^{-6}$ and continued from the previous round’s checkpoint during adversarial training.

\paragraph{Baselines}

We compare RADAR against both prompt-only and trainable baselines. For prompt-only baselines, we use the same three instruction-tuned LLM backbones as zero-shot detectors, without any fine-tuning. For trainable baselines, we use the same DeBERTa-v3-base detector backbone as RADAR to ensure a controlled comparison.

We consider two training-based baselines: (1) \textit{Standard Supervised Training}, where a fixed generator produces 2,500 fake articles, matching the total number of fake training examples generated across RADAR's five adversarial rounds, and the detector is trained once on the resulting static training set; and (2) \textit{Static Adversarial Training}, where a fixed generator produces 500 fake articles per round over five rounds, matching RADAR's round-wise data volume, and the detector is updated round by round on the corresponding static adversarial data, but without generator adaptation. For all methods, we report ROC-AUC under two detector settings: \textit{w/o Retrieval} (D$-$) and \textit{w/ Retrieval} (D$+$), where the latter augments the detector with retrieved evidence from news articles.

\begin{table*}[t]
    \centering
    \small
    \renewcommand{\arraystretch}{1.1}
    \begin{tabular*}{\textwidth}{@{\extracolsep{\fill}}p{4cm}lcc@{}}
        \toprule
        \textbf{Baseline / Setting} & \textbf{Model / Configuration} & \textbf{w/o Retrieval (D$-$)} & \textbf{w/ Retrieval (D$+$)} \\
        \midrule
        
        \multirow{3}{4cm}{\emph{Zero-shot LLM Baselines}} 
        & Qwen3-4B-Instruct   & 55.73 & 59.43 \\
        & Gemma3-4B           & 49.40 & 55.94 \\
        & Phi-3.5-mini (3.8B) & 42.01 & 68.39 \\
        
        \midrule
        \multirow{3}{4cm}{\emph{Trainable Detector vs. Generator: Qwen3-4B-Instruct}} 
        & Standard Supervised Training & 57.41 & 67.64 \\
        & Static Adversarial Training  & 55.44 & 71.36 \\
        & \textbf{RADAR}               & \textbf{61.65} & \textbf{74.18} \\
        
        \midrule
        \multirow{3}{4cm}{\emph{Trainable Detector vs. Generator: Gemma3-4B}} 
        & Standard Supervised Training & 58.21 & 71.29 \\
        & Static Adversarial Training  & 51.13 & 68.30 \\
        & \textbf{RADAR}               & \textbf{59.67} & \textbf{75.13} \\
        
        \midrule
        \multirow{3}{4cm}{\emph{Trainable Detector vs. Generator: Phi-3.5-mini (3.8B)}} 
        & Standard Supervised Training & 58.59 & 64.10 \\
        & Static Adversarial Training  & 59.94 & 69.90 \\
        & \textbf{RADAR}               & \textbf{68.43} & \textbf{75.58} \\
        
        \bottomrule
    \end{tabular*}
    \caption{Main results on the \textit{AdvFake-News-Please} evaluation set (ROC-AUC, \%). We compare zero-shot LLM detectors with trained DeBERTa-v3-base baselines under three training settings and the full RADAR framework.} 
    \label{tab:main_results}
\end{table*}

\begin{figure*}[t]
\centering
\begin{subfigure}{0.32\textwidth}
\centering
\includegraphics[width=\linewidth]{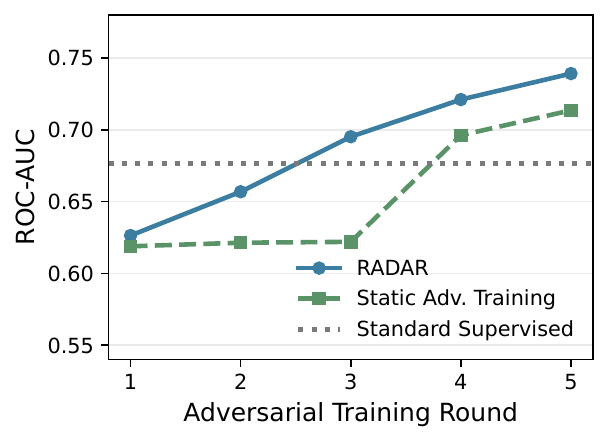}
\caption{Qwen3-4B-Instruct}
\end{subfigure}
\hfill
\begin{subfigure}{0.32\textwidth}
\centering
\includegraphics[width=\linewidth]{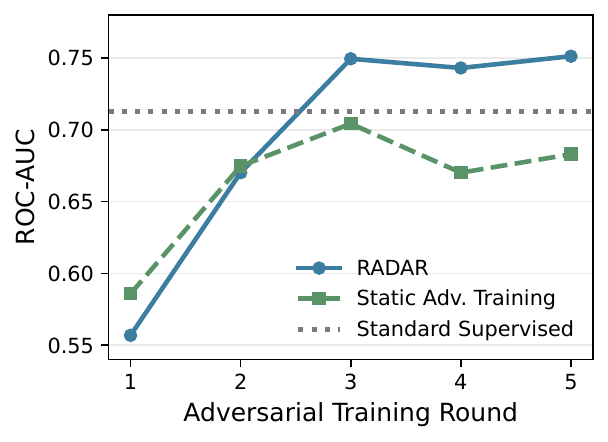}
\caption{Gemma3-4B}
\end{subfigure}
\hfill
\begin{subfigure}{0.32\textwidth}
\centering
\includegraphics[width=\linewidth]{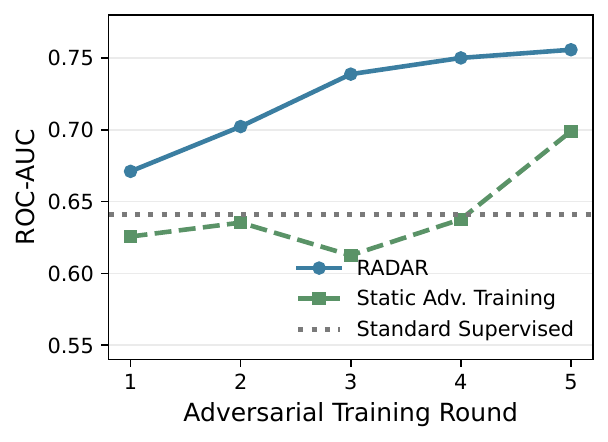}
\caption{Phi-3.5-mini}
\end{subfigure}
\caption{Round-wise detector ROC-AUC across adversarial training rounds. RADAR generally improves over rounds and reaches higher final performance than Static Adversarial Training, while Standard Supervised Training is shown as a fixed reference line.}
\label{fig:roundwise_auc}
\end{figure*}

\begin{table*}[t]
    \small
    \centering
    \setlength{\tabcolsep}{0pt}
    \begin{tabular*}{\textwidth}{@{\extracolsep{\fill}}lccc@{}}
        \toprule
        \textbf{Configuration} & \textbf{Qwen3-4B-Instruct} & \textbf{Gemma3-4B} & \textbf{Phi-3.5-mini} \\
        \midrule
        No Retrieval (G$-$ / D$-$) & 59.41 & 53.65 & 65.91 \\
        Retrieval-Augmented Generator Only (G$+$ / D$-$) & 61.65 & 59.67 & 68.43 \\
        Retrieval-Augmented Detector Only (G$-$ / D$+$) & 71.94 & 65.35 & 71.99 \\
        Retrieval-Augmented Generator \& Detector (G$+$ / D$+$) & \textbf{74.18} & \textbf{75.13} & \textbf{75.58} \\
        \bottomrule
    \end{tabular*}
    \caption{Impact of retrieval configuration across generator backbones using the same detector backbone. G denotes generator-side retrieval and D denotes detector-side retrieval. We report detector performance in ROC-AUC (\%).}
    \label{tab:rag_comparison}
\end{table*}

\subsection{Main Results}

Table~\ref{tab:main_results} compares RADAR with prompt-only LLM detectors and trainable DeBERTa baselines under both D$-$ and D$+$ settings.

Overall, prompt-only LLM detectors show limited performance in this benchmark, particularly without retrieved evidence. Adding retrieval improves these baselines, but the best prompt-only result still remains below the full RADAR setting.

Across all generator backbones, RADAR achieves the strongest performance among trainable detectors under both D$-$ and D$+$ settings. Compared with the two trainable baselines, RADAR consistently performs better under matched detector backbones and retrieval settings. This suggests that the improvement is not simply due to using a trainable DeBERTa detector or increasing the amount of generated training data. Rather, the adaptive refinement loop provides more informative adversarial examples for detector training, leading to stronger final performance.

\subsection{Effect of Iterative Adversarial Refinement}

To examine whether RADAR benefits from iterative adversarial refinement rather than only from additional adversarial data, we track detector ROC-AUC after each training round and compare RADAR with Static Adversarial Training and Standard Supervised Training. As shown in \autoref{fig:roundwise_auc}, RADAR generally improves over successive rounds across all three generator backbones, while Static Adversarial Training is less stable and often plateaus at a lower level. This suggests that adapting the generator with feedback and cached successful attacks exposes the detector to progressively more informative training examples, yielding stronger final performance than round-wise detector updates with a fixed attacker.

\subsection{Impact of RAG Components}

To isolate the contribution of retrieval on each side, we compare four retrieval configurations across three generator backbones in \autoref{tab:rag_comparison}. Enabling retrieval on either side generally improves performance over the no-retrieval setting, and the strongest results are obtained when both generator-side and detector-side retrieval are enabled.

These results suggest that the two retrieval roles are complementary. Detector-side retrieval provides external evidence for classification, while generator-side retrieval supplies realism priors for constructing fact-altered rewrites. Their combination yields the most effective adversarial training setup across all three generator backbones.

\definecolor{lightred}{RGB}{255,220,220}
\definecolor{lightgreen}{RGB}{220,255,220}

\newcommand{\chg}[1]{{\sethlcolor{lightred}\hl{#1}}}
\newcommand{\add}[1]{{\sethlcolor{lightgreen}\hl{#1}}}

\begin{table}[t]
\centering
\small
\setlength{\tabcolsep}{10pt}
\renewcommand{\arraystretch}{1.15}

% box padding / border
\setlength{\fboxsep}{5pt}
\setlength{\fboxrule}{0.6pt}

% --- Top: narrower boxed text block ---
\fbox{%
  \begin{minipage}{0.92\linewidth}
  \small
  \setlength{\baselineskip}{13pt}
  \textbf{Before rewrite:}\\
  \chg{The local commuters} inform \chg{the Gulf News} that these \chg{adjustments} will affect their \chg{budgets} and they might need to \chg{budget} their \chg{expenditures} more \chg{diligently}.\\[6pt]
  \textbf{After rewrite:}\\
  \add{Motorists} inform \add{Khaleej Times} that these \add{modifications} will affect their \add{finances} and they might need to \add{handle} their \add{spending} more \add{prudently}.
  \end{minipage}%
}

\vspace{6pt}

\begin{tabular*}{0.95\linewidth}{@{\extracolsep{\fill}}l r l r@{}}
\toprule
\multicolumn{2}{c}{\textbf{Before (Top-5 IG)}} &
\multicolumn{2}{c}{\textbf{After (Top-5 IG)}} \\
\midrule
diligently   & 3.02 & spending      & 5.67 \\
commuters    & 2.94 & modifications & 4.81 \\
expenditures & 2.62 & prudently     & 4.67 \\
Gulf         & 2.50 & handle        & 4.22 \\
local        & 2.22 & finances      & 4.10 \\
\bottomrule
\end{tabular*}
\vspace{-1mm}
\caption{Token-salience visualization using Integrated Gradients. Salient tokens in the original sentence are highlighted in red, and their rewritten counterparts in the generated sentence are highlighted in green.}
\label{tab:token_salience_example}
\vspace{-2mm}
\end{table}

\subsection{Qualitative Analysis of Token Salience}
To illustrate how the detector attends to suspicious spans, we visualize token-level salience using Integrated Gradients on a rewrite pair. As shown in \autoref{tab:token_salience_example}, highly salient tokens concentrate on source-related entities and budget-related claim words, which are precisely the anchors replaced by the generator during rewriting.

\begin{table}[t]
\centering
\small
\begin{tabular}{lccc}
\toprule
 & \textbf{Qwen3} & \textbf{Gemma3} & \textbf{Phi-3.5} \\
 %& \textbf{ROC-AUC} & \textbf{ROC-AUC} & \textbf{AUC} \\
\midrule
RADAR & \bf 74.18 & \bf 75.13 & \bf 75.58 \\
w/o VAF & 71.22 & 73.55 & 72.61 \\
w/o Few-shot & 70.80 & 72.87 & 71.56 \\
w/o VAF \& Few-shot & 66.51 & 72.49 & 71.19 \\
\bottomrule
\end{tabular}
\caption{Ablation of VAF and few-shot feedback components under the dual-retrieval setting (G$+$ / D$+$) across different generators. We report detector ROC-AUC (\%).}
\label{tab:feedback_ablation}
\vspace{-2mm}
\end{table}

\iffalse
\begin{table}[t]
\centering
\small
%\setlength{\tabcolsep}{1pt}
\resizebox{\columnwidth}{!}{
\begin{tabular}{lcccccc}
\toprule
& \multicolumn{2}{c}{\textbf{Qwen3-4B-Instruct}} & \multicolumn{2}{c}{\textbf{Gemma3-4B}} & \multicolumn{2}{c}{\textbf{Phi-3.5-mini}} \\
\cmidrule(lr){2-3} \cmidrule(lr){4-5} \cmidrule(lr){6-7}
 & \textbf{AUC} & $\Delta$ & \textbf{AUC} & $\Delta$ & \textbf{AUC} & $\Delta$ \\
\midrule
RADAR & 74.18 & -- & 75.13 & -- & 75.58 & -- \\
w/o VAF & 71.22 & -2.96 & 73.55 & -1.58 & 72.61 & -2.97 \\
w/o Few-shot & 70.80 & -3.38 & 72.87 & -2.26 & 71.56 & -4.02 \\
w/o Both & 66.51 & -7.67 & 72.49 & -2.64 & 71.19 & -4.39 \\
\bottomrule
\end{tabular}}
\caption{Ablation of VAF and few-shot feedback components under the dual-retrieval setting (G$+$ / D$+$) across different generator backbones. We report detector ROC-AUC (\%), and $\Delta$ is computed relative to full RADAR for each generator.}
\label{tab:feedback_ablation}
\vspace{-2mm}
\end{table}
\fi

\begin{table}[t]
    \centering
    \small
    \setlength{\tabcolsep}{3pt}
    \renewcommand{\arraystretch}{1.1}
    \begin{tabular}{lcc}
        \toprule
        \textbf{Model / Configuration} & \textbf{w/o Retrieval} & \textbf{w/ Retrieval} \\
        \midrule
        
        \multicolumn{3}{l}{\emph{Zero-shot LLM Baselines}} \\
        %\midrule
        Qwen3-4B-Instruct   & 46.02 & 52.22 \\
        Gemma3-4B           & 48.48 & 52.10 \\
        Phi-3.5-mini (3.8B) & 41.05 & 60.43 \\
        
        \midrule
        \multicolumn{3}{l}{\emph{Trainable Detector vs. Generator: Qwen3-4B-Instruct}} \\
        %\midrule
        Standard Supervised Training & 58.52 & 58.76 \\
        Static Adversarial Training  & 52.37 & 63.88 \\
        \textbf{RADAR}               & \textbf{60.56} & \textbf{69.20} \\
        
        \midrule
        \multicolumn{3}{l}{\emph{Trainable Detector vs. Generator: Gemma3-4B}} \\
        %\midrule
        Standard Supervised Training & 57.92 & 65.94 \\
        Static Adversarial Training  & 52.37 & 63.15 \\
        \textbf{RADAR}               & \textbf{60.28} & \textbf{69.35} \\
        
        \midrule
        \multicolumn{3}{l}{\emph{Trainable Detector vs. Generator: Phi-3.5-mini (3.8B)}} \\
        %\midrule
        Standard Supervised Training & 58.74 & 64.90 \\
        Static Adversarial Training  & 54.28 & 66.52 \\
        \textbf{RADAR}               & \textbf{61.21} & \textbf{69.59} \\
        
        \bottomrule
    \end{tabular}
    \caption{Cross-attacker transfer results on fake news generated by GPT-5.3 Instant, an unseen external attacker used only at evaluation time. We report ROC-AUC (\%) under w/o Retrieval (D$-$) and w/ Retrieval (D$+$).}
    \label{tab:cross_attacker_transfer}
\end{table}

\subsection{Effect of Feedback Design}

We ablate the two feedback components in RADAR under the dual-retrieval setting (G$+$ / D$+$) across three generator backbones. LoRA-based generator adaptation remains enabled in all configurations, so the ablation isolates the effects of VAF and few-shot demonstrations while keeping parameter-efficient generator updates fixed. As shown in \autoref{tab:feedback_ablation}, removing either VAF or few-shot demonstrations consistently hurts performance, while removing both causes the largest drop in every setting.

This suggests that structured detector feedback and successful-attack examples provide complementary signals for generator refinement, with their combination yielding the strongest results across all three generator backbones.

\subsection{Cross-Attacker Transferability}

To verify that RADAR does not merely overfit to the specific fake-news distribution in the \textit{AdvFake-News-Please} evaluation set, we construct an additional evaluation setting using \textbf{GPT-5.3 Instant} \cite{openai2026gpt53instant} as an unseen external attacker. We evaluate all methods on fake news generated by this external LLM, which is never used during training, to test transferability to unseen attacks.

Table~\ref{tab:cross_attacker_transfer} reports the results. Under this external-attacker setting, zero-shot LLM detectors remain limited, indicating that unseen LLM-generated attacks are challenging without task-specific adaptation. In contrast, across all three generator settings, RADAR consistently outperforms both \textit{Standard Supervised Training} and \textit{Static Adversarial Training} under the same detector backbone. These results suggest that iterative co-evolution improves robustness to attacks from an unseen external generator beyond the original co-evolutionary setting.

\section{Conclusion}

We proposed RADAR, a retrieval-augmented adversarial training framework for adaptive LLM-generated fake-news detection. RADAR combines a trainable encoder-based detector with an adaptive generator, dual-side retrieval, and Verbal Adversarial Feedback (VAF) to study how retrieval and structured feedback affect detector training under rewrite-style attacks. Experiments on \textit{AdvFake-News-Please} show that RADAR outperforms prompt-only LLM detectors and trainable static baselines, with dual-side retrieval, VAF, and few-shot examples offering complementary benefits. Finally, transfer evaluation with an unseen external attacker shows that RADAR improves cross-attacker transferability. Overall, RADAR demonstrates that evidence grounding and interpretable adversarial refinement are a promising combination for adaptive fake news detection.

\section*{Limitations}

Although RADAR shows positive cross-attacker transfer to an unseen external generator, our empirical study is still centered on a single benchmark, \textit{AdvFake-News-Please}. This choice is intentional rather than incidental. RADAR is designed to study retrieval-augmented fake-news detection in a leakage-aware setting, where retrieval should provide non-trivial external evidence rather than simply duplicate knowledge already memorized by the detector backbone. Since our detector, DeBERTa-v3-base, is pretrained largely on data ending around early 2021, we prioritize post-2021 news to reduce potential overlap between parametric knowledge and evaluation content. Many widely used fake-news benchmarks, such as FakeNewsNet and BuzzFeed, are dominated by earlier news and may therefore under-test the marginal value of retrieval in our setting. In addition, RADAR requires a dataset that jointly provides text-only real/fake news pairs and a compatible news corpus for evidence retrieval within a similar time range; we found very few public benchmarks satisfying these constraints. As a result, our current evaluation remains limited to a single dataset, and broader cross-dataset validation remains an important direction for future work.

Our evaluation also points to several directions for future work. First, given the computational cost of round-wise generator–detector co-evolution, future studies could examine robustness across multiple random seeds and quantify run-to-run variance. Second, complementary human evaluation could further assess the plausibility, factual alterations, and strategy compliance of generated adversarial examples, providing a more comprehensive characterization of generation quality. Finally, while our lightweight encoder-based detector supports efficient iterative updates, extending RADAR to trainable generative reasoning LLM detectors would help assess its applicability across different detector architectures. Future work could also benchmark inference latency and computational cost to better characterize the accuracy–efficiency trade-off.

\section*{Ethical Considerations}
Our framework trains a generator to produce realistic, fact-altered rewrites and iteratively improves it with detector-provided verbal adversarial feedback. While this serves the defensive goal of strengthening fake-news detection, it also introduces a clear dual-use risk: the same mechanisms (retrieval-guided realism priors and targeted feedback) could be repurposed to craft more deceptive misinformation that better evades existing detectors.

To mitigate potential misuse, we (i) position RADAR strictly as a \emph{defensive training} pipeline and do not advocate deploying the generator for content creation, (ii) encourage safeguards such as rate limiting, logging, and human-in-the-loop review for any red-team use, and (iii) support complementary ecosystem measures when deploying detection systems. Finally, we emphasize that our experiments are conducted on public benchmarks and aim to improve robustness against adaptive attacks, not to enable real-world disinformation campaigns.

\bibliography{custom}

@article{douze2024faiss,
  title   = {The Faiss Library},
  author  = {Douze, Matthijs and Guzhva, Alexandr and Deng, Chengqi and Johnson, Jeff and Szilvasy, Gergely and Mazar{\'e}, Pierre-Emmanuel and Lomeli, Maria and Hosseini, Lucas and J{\'e}gou, Herv{\'e}},
  journal = {arXiv preprint arXiv:2401.08281},
  year    = {2024}
}

@misc{openai2026gpt53instant,
  author       = {{OpenAI}},
  title        = {{GPT-5.3 Instant: Smoother, more useful everyday conversations}},
  year         = {2026},
  month        = mar,
  howpublished = {\url{https://openai.com/index/gpt-5-3-instant/}},
  note         = {Accessed: 2026-03-14}
}

@article{yang2025qwen3,
  title        = {Qwen3 Technical Report},
  author       = {Yang, An and Li, Anfeng and Yang, Baosong and Zhang, Beichen and Hui, Binyuan and Zheng, Bo and Yu, Bowen and Gao, Chang and Huang, Chengen and Lv, Chenxu and Zheng, Chujie and Liu, Dayiheng and Zhou, Fan and Huang, Fei and Hu, Feng and Ge, Hao and Wei, Haoran and Lin, Huan and Tang, Jialong and Yang, Jian and Tu, Jianhong and Zhang, Jianwei and Yang, Jianxin and Yang, Jiaxi and Zhou, Jing and Zhou, Jingren and Lin, Junyang and Dang, Kai and Bao, Keqin and Yang, Kexin and Yu, Le and Deng, Lianghao and Li, Mei and Xue, Mingfeng and Li, Mingze and Zhang, Pei and Wang, Peng and Zhu, Qin and Men, Rui and Gao, Ruize and Liu, Shixuan and Luo, Shuang and Li, Tianhao and Tang, Tianyi and Yin, Wenbiao and Ren, Xingzhang and Wang, Xinyu and Zhang, Xinyu and Ren, Xuancheng and Fan, Yang and Su, Yang and Zhang, Yichang and Zhang, Yinger and Wan, Yu and Liu, Yuqiong and Wang, Zekun and Cui, Zeyu and Zhang, Zhenru and Zhou, Zhipeng and Qiu, Zihan},
  journal      = {arXiv preprint arXiv:2505.09388},
  year         = {2025}
}

@article{gteam2025gemma3,
  title        = {Gemma 3 Technical Report},
  author       = {{Gemma Team}},
  journal      = {arXiv preprint arXiv:2503.19786},
  year         = {2025}
}

@misc{abdin2024phi3,
      title={Phi-3 Technical Report: A Highly Capable Language Model Locally on Your Phone}, 
      author={Marah Abdin and Jyoti Aneja and Hany Awadalla and Ahmed Awadallah and Ammar Ahmad Awan and Nguyen Bach and Amit Bahree and Arash Bakhtiari and Jianmin Bao and Harkirat Behl and Alon Benhaim and Misha Bilenko and Johan Bjorck and Sébastien Bubeck and Martin Cai and Qin Cai and Vishrav Chaudhary and Dong Chen and Dongdong Chen and Weizhu Chen and Yen-Chun Chen and Yi-Ling Chen and Hao Cheng and Parul Chopra and Xiyang Dai and Matthew Dixon and Ronen Eldan and Victor Fragoso and Jianfeng Gao and Mei Gao and Min Gao and Amit Garg and Allie Del Giorno and Abhishek Goswami and Suriya Gunasekar and Emman Haider and Junheng Hao and Russell J. Hewett and Wenxiang Hu and Jamie Huynh and Dan Iter and Sam Ade Jacobs and Mojan Javaheripi and Xin Jin and Nikos Karampatziakis and Piero Kauffmann and Mahoud Khademi and Dongwoo Kim and Young Jin Kim and Lev Kurilenko and James R. Lee and Yin Tat Lee and Yuanzhi Li and Yunsheng Li and Chen Liang and Lars Liden and Xihui Lin and Zeqi Lin and Ce Liu and Liyuan Liu and Mengchen Liu and Weishung Liu and Xiaodong Liu and Chong Luo and Piyush Madan and Ali Mahmoudzadeh and David Majercak and Matt Mazzola and Caio César Teodoro Mendes and Arindam Mitra and Hardik Modi and Anh Nguyen and Brandon Norick and Barun Patra and Daniel Perez-Becker and Thomas Portet and Reid Pryzant and Heyang Qin and Marko Radmilac and Liliang Ren and Gustavo de Rosa and Corby Rosset and Sambudha Roy and Olatunji Ruwase and Olli Saarikivi and Amin Saied and Adil Salim and Michael Santacroce and Shital Shah and Ning Shang and Hiteshi Sharma and Yelong Shen and Swadheen Shukla and Xia Song and Masahiro Tanaka and Andrea Tupini and Praneetha Vaddamanu and Chunyu Wang and Guanhua Wang and Lijuan Wang and Shuohang Wang and Xin Wang and Yu Wang and Rachel Ward and Wen Wen and Philipp Witte and Haiping Wu and Xiaoxia Wu and Michael Wyatt and Bin Xiao and Can Xu and Jiahang Xu and Weijian Xu and Jilong Xue and Sonali Yadav and Fan Yang and Jianwei Yang and Yifan Yang and Ziyi Yang and Donghan Yu and Lu Yuan and Chenruidong Zhang and Cyril Zhang and Jianwen Zhang and Li Lyna Zhang and Yi Zhang and Yue Zhang and Yunan Zhang and Xiren Zhou},
      year={2024},
      eprint={2404.14219},
      archivePrefix={arXiv},
      primaryClass={cs.CL},
      url={https://arxiv.org/abs/2404.14219}, 
}

@inproceedings{tian-etal-2025-symbolic,
    title = "A Symbolic Adversarial Learning Framework for Evolving Fake News Generation and Detection",
    author = "Tian, Chong  and
      Ho, Qirong  and
      Chen, Xiuying",
    editor = "Christodoulopoulos, Christos  and
      Chakraborty, Tanmoy  and
      Rose, Carolyn  and
      Peng, Violet",
    booktitle = "Proceedings of the 2025 Conference on Empirical Methods in Natural Language Processing",
    month = nov,
    year = "2025",
    address = "Suzhou, China",
    publisher = "Association for Computational Linguistics",
    url = "https://aclanthology.org/2025.emnlp-main.619/",
    doi = "10.18653/v1/2025.emnlp-main.619",
    pages = "12296--12310",
    ISBN = "979-8-89176-332-6",
}

@inproceedings{niu-etal-2024-veract,
  title     = {VeraCT Scan: Retrieval-Augmented Fake News Detection with Justifiable Reasoning},
  author    = {Niu, Cheng and Guan, Yang and Wu, Yuanhao and Zhu, Juno and Song, Juntong and Zhong, Randy and Zhu, Kaihua and Xu, Siliang and Diao, Shizhe and Zhang, Tong},
  booktitle = {Proceedings of the 62nd Annual Meeting of the Association for Computational Linguistics (Volume 3: System Demonstrations)},
  year      = {2024},
  publisher = {Association for Computational Linguistics},
  pages     = {266--277},
  doi       = {10.18653/v1/2024.acl-demos.25},
  url       = {https://aclanthology.org/2024.acl-demos.25/}
}

@inproceedings{singh-namin-2024-adversarial-rag-fake-news,
  title     = {Adversarial Training of Retrieval Augmented Generation to Generate Believable Fake News},
  author    = {Singh, Sonali and Namin, Akbar Siami},
  booktitle = {{IEEE} International Conference on Big Data, BigData 2024, Washington, DC, USA, December 15--18, 2024},
  year      = {2024},
  publisher = {{IEEE}},
  pages     = {3589--3598},
  doi       = {10.1109/BIGDATA62323.2024.10825933},
  url       = {https://doi.org/10.1109/BigData62323.2024.10825933}
}

@inproceedings{chen-etal-2025-real,
  title     = {Real-time Factuality Assessment from Adversarial Feedback},
  author    = {Chen, Sanxing and Huang, Yukun and Dhingra, Bhuwan},
  booktitle = {Proceedings of the 63rd Annual Meeting of the Association for Computational Linguistics (Volume 1: Long Papers)},
  year      = {2025},
  publisher = {Association for Computational Linguistics},
  pages     = {1610--1630},
  doi       = {10.18653/v1/2025.acl-long.81},
  url       = {https://aclanthology.org/2025.acl-long.81/}
}

@inproceedings{zellers-etal-2019-defending,
  title     = {Defending Against Neural Fake News},
  author    = {Zellers, Rowan and Holtzman, Ari and Rashkin, Hannah and Bisk, Yonatan and Farhadi, Ali and Roesner, Franziska and Choi, Yejin},
  booktitle = {Advances in Neural Information Processing Systems 32: Annual Conference on Neural Information Processing Systems 2019 (NeurIPS 2019)},
  year      = {2019},
  pages     = {9051--9062},
  url       = {https://proceedings.neurips.cc/paper/2019/hash/3e9f0fc9b2f89e043bc6233994dfcf76-Abstract.html}
}

@inproceedings{wang-etal-2025-llm-gan,
  title     = {{LLM-GAN:} Constructing Generative Adversarial Network Through Large Language Models for Explainable Fake News Detection},
  author    = {Wang, Yifeng and Gu, Zhouhong and Zhang, Siwei and Zheng, Suhang and Wang, Tao and Li, Tianyu and Feng, Hongwei and Xiao, Yanghua},
  booktitle = {2025 {IEEE} International Conference on Acoustics, Speech and Signal Processing, {ICASSP} 2025, Hyderabad, India, April 6--11, 2025},
  year      = {2025},
  publisher = {{IEEE}},
  pages     = {1--5},
  doi       = {10.1109/ICASSP49660.2025.10889048},
  url       = {https://doi.org/10.1109/ICASSP49660.2025.10889048}
}

@inproceedings{nezafat-samet-2024-rag-genai-fake-news,
  title     = {Fake News Detection with Retrieval Augmented Generative Artificial Intelligence},
  author    = {Nezafat, Mohammad Vatani and Samet, Saeed},
  booktitle = {2nd International Conference on Foundation and Large Language Models, {FLLM} 2024, Dubai, United Arab Emirates, November 26--29, 2024},
  year      = {2024},
  publisher = {{IEEE}},
  pages     = {160--167},
  doi       = {10.1109/FLLM63129.2024.10852474},
  url       = {https://doi.org/10.1109/FLLM63129.2024.10852474}
}

@article{
doi:10.1126/science.aap9559,
author = {Soroush Vosoughi  and Deb Roy  and Sinan Aral },
title = {The spread of true and false news online},
journal = {Science},
volume = {359},
number = {6380},
pages = {1146-1151},
year = {2018},
doi = {10.1126/science.aap9559},
URL = {https://www.science.org/doi/abs/10.1126/science.aap9559}}

@inproceedings{he2021debertadecodingenhancedbertdisentangled,
  title     = {DeBERTa: Decoding-Enhanced BERT with Disentangled Attention},
  author    = {He, Pengcheng and Liu, Xiaodong and Gao, Jianfeng and Chen, Weizhu},
  booktitle = {International Conference on Learning Representations (ICLR)},
  year      = {2021},
  url       = {https://openreview.net/forum?id=XPZIaotutsD}
}

@inproceedings{jin2020bertreallyrobuststrong,
  title     = {Is BERT Really Robust? A Strong Baseline for Natural Language Attack on Text Classification and Entailment},
  author    = {Jin, Di and Jin, Zhijing and Zhou, Joey Tianyi and Szolovits, Peter},
  booktitle = {Proceedings of the AAAI Conference on Artificial Intelligence},
  year      = {2020},
  pages     = {8018--8025},
  doi       = {10.1609/AAAI.V34I05.6311},
  url       = {https://dblp.org/rec/conf/aaai/JinJZS20}
}

@inproceedings{yu2017seqgansequencegenerativeadversarial,
  title     = {SeqGAN: Sequence Generative Adversarial Nets with Policy Gradient},
  author    = {Yu, Lantao and Zhang, Weinan and Wang, Jun and Yu, Yong},
  booktitle = {Proceedings of the AAAI Conference on Artificial Intelligence},
  year      = {2017},
  pages     = {2852--2858},
  doi       = {10.1609/aaai.v31i1.10804},
  url       = {https://ojs.aaai.org/index.php/AAAI/article/view/10804}
}

@inproceedings{caccia2020languagegansfallingshort,
  title     = {Language GANs Falling Short},
  author    = {Caccia, Massimo and Caccia, Lucas and Fedus, William and Larochelle, Hugo and Pineau, Joelle and Charlin, Laurent},
  booktitle = {International Conference on Learning Representations (ICLR)},
  year      = {2020},
  url       = {https://openreview.net/forum?id=BJgza6VtPB}
}

@inproceedings{madaan2023selfrefineiterativerefinementselffeedback,
  title     = {Self-Refine: Iterative Refinement with Self-Feedback},
  author    = {Madaan, Aman and Tandon, Niket and Gupta, Prakhar and Hallinan, Skyler and Gao, Luyu and Wiegreffe, Sarah and Alon, Uri and Dziri, Nouha and Prabhumoye, Shrimai and Yang, Yiming and Gupta, Shashank and Majumder, Bodhisattwa Prasad and Hermann, Katherine and Welleck, Sean and Yazdanbakhsh, Amir and Clark, Peter},
  booktitle = {Advances in Neural Information Processing Systems (NeurIPS)},
  year      = {2023},
  url       = {https://papers.nips.cc/paper_files/paper/2023/hash/91edff07232fb1b55a505a9e9f6c0ff3-Abstract-Conference.html}
}

@inproceedings{shinn2023reflexionlanguageagentsverbal,
  title     = {Reflexion: Language Agents with Verbal Reinforcement Learning},
  author    = {Shinn, Noah and Cassano, Federico and Gopinath, Ashwin and Narasimhan, Karthik and Yao, Shunyu},
  booktitle = {Advances in Neural Information Processing Systems (NeurIPS)},
  year      = {2023},
  url       = {https://papers.nips.cc/paper_files/paper/2023/hash/1b44b878bb782e6954cd888628510e90-Abstract-Conference.html}
}

@inproceedings{goodfellow2014generativeadversarialnetworks,
  title     = {Generative Adversarial Nets},
  author    = {Goodfellow, Ian J. and Pouget-Abadie, Jean and Mirza, Mehdi and Xu, Bing and Warde-Farley, David and Ozair, Sherjil and Courville, Aaron C. and Bengio, Yoshua},
  booktitle = {Advances in Neural Information Processing Systems (NeurIPS)},
  year      = {2014},
  pages     = {2672--2680},
  url       = {https://papers.nips.cc/paper/5423-generative-adversarial-nets}
}

@inproceedings{guo2017longtextgenerationadversarial,
  title     = {Long Text Generation via Adversarial Training with Leaked Information},
  author    = {Guo, Jiaxian and Lu, Sidi and Cai, Han and Zhang, Weinan and Yu, Yong and Wang, Jun},
  booktitle = {Proceedings of the Thirty-Second AAAI Conference on Artificial Intelligence (AAAI-18)},
  year      = {2018},
  pages     = {5141--5148},
  doi       = {10.1609/aaai.v32i1.11957},
  url       = {https://ojs.aaai.org/index.php/AAAI/article/view/11957}
}

@inproceedings{lin2018adversarialrankinglanguagegeneration,
  title     = {Adversarial Ranking for Language Generation},
  author    = {Lin, Kevin and Li, Dianqi and He, Xiaodong and Zhang, Zhengyou and Sun, Ming-Ting},
  booktitle = {Advances in Neural Information Processing Systems 30},
  year      = {2017},
  pages     = {3155--3165},
  url       = {https://papers.nips.cc/paper/6908-adversarial-ranking-for-language-generation}
}

@inproceedings{wang2017liarliarpantsfire,
  title     = {{``Liar, Liar Pants on Fire''}: A New Benchmark Dataset for Fake News Detection},
  author    = {Wang, William Yang},
  booktitle = {Proceedings of the 55th Annual Meeting of the Association for Computational Linguistics (Volume 2: Short Papers)},
  year      = {2017},
  pages     = {422--426},
  doi       = {10.18653/v1/P17-2067},
  url       = {https://aclanthology.org/P17-2067/}
}

@article{10.1007/s11042-020-10183-2,
author = {Kaliyar, Rohit Kumar and Goswami, Anurag and Narang, Pratik},
title = {FakeBERT: Fake news detection in social media with a BERT-based deep learning approach},
year = {2021},
publisher = {Kluwer Academic Publishers},
volume = {80},
number = {8},
url = {https://doi.org/10.1007/s11042-020-10183-2},
doi = {10.1007/s11042-020-10183-2},
journal = {Multimedia Tools Appl.},
pages = {11765–11788}
}

@inproceedings{lewis2021retrievalaugmentedgenerationknowledgeintensivenlp,
  title     = {Retrieval-Augmented Generation for Knowledge-Intensive NLP Tasks},
  author    = {Lewis, Patrick and Perez, Ethan and Piktus, Aleksandra and Petroni, Fabio and Karpukhin, Vladimir and Goyal, Naman and K{\"u}ttler, Heinrich and Lewis, Mike and Yih, Wen-tau and Rockt{\"a}schel, Tim and Riedel, Sebastian and Kiela, Douwe},
  booktitle = {Advances in Neural Information Processing Systems 33},
  year      = {2020},
  pages     = {9459--9474},
  url       = {https://proceedings.neurips.cc/paper/2020/hash/6b493230205f780e1bc26945df7481e5-Abstract.html}
}

@misc{bai2022constitutionalaiharmlessnessai,
      title={Constitutional AI: Harmlessness from AI Feedback}, 
      author={Yuntao Bai and Saurav Kadavath and Sandipan Kundu and Amanda Askell and Jackson Kernion and Andy Jones and Anna Chen and Anna Goldie and Azalia Mirhoseini and Cameron McKinnon and Carol Chen and Catherine Olsson and Christopher Olah and Danny Hernandez and Dawn Drain and Deep Ganguli and Dustin Li and Eli Tran-Johnson and Ethan Perez and Jamie Kerr and Jared Mueller and Jeffrey Ladish and Joshua Landau and Kamal Ndousse and Kamile Lukosuite and Liane Lovitt and Michael Sellitto and Nelson Elhage and Nicholas Schiefer and Noemi Mercado and Nova DasSarma and Robert Lasenby and Robin Larson and Sam Ringer and Scott Johnston and Shauna Kravec and Sheer El Showk and Stanislav Fort and Tamera Lanham and Timothy Telleen-Lawton and Tom Conerly and Tom Henighan and Tristan Hume and Samuel R. Bowman and Zac Hatfield-Dodds and Ben Mann and Dario Amodei and Nicholas Joseph and Sam McCandlish and Tom Brown and Jared Kaplan},
      year={2022},
      eprint={2212.08073},
      archivePrefix={arXiv},
      primaryClass={cs.CL},
      url={https://arxiv.org/abs/2212.08073}, 
}

@inproceedings{karpukhin2020densepassageretrievalopendomain,
  title     = {Dense Passage Retrieval for Open-Domain Question Answering},
  author    = {Karpukhin, Vladimir and O{\u{g}}uz, Barlas and Min, Sewon and Lewis, Patrick and Wu, Ledell and Edunov, Sergey and Chen, Danqi and Yih, Wen-tau},
  booktitle = {Proceedings of the 2020 Conference on Empirical Methods in Natural Language Processing (EMNLP)},
  year      = {2020},
  pages     = {6769--6781},
  doi       = {10.18653/v1/2020.emnlp-main.550},
  url       = {https://aclanthology.org/2020.emnlp-main.550/}
}

@article{yuksekgonul2024textgradautomaticdifferentiationtext,
  title   = {Optimizing generative AI by backpropagating language model feedback},
  author  = {Yuksekgonul, Mert and Bianchi, Federico and Boen, Joseph and Liu, Sheng and Lu, Pan and Huang, Zhi and Guestrin, Carlos and Zou, James},
  journal = {Nature},
  year    = {2025},
  volume  = {639},
  number  = {8055},
  pages   = {609--616},
  doi     = {10.1038/s41586-025-08661-4},
  url     = {https://doi.org/10.1038/s41586-025-08661-4}
}

\appendix

\section{Additional Analyses}
\label{app:additional_analysis}

\subsection{Retrieval Overlap Analysis}
\label{app:retrieval_overlap}

A potential concern is that the gains from retrieval may partly reflect a trivial ``lookup advantage,'' where the retriever returns passages that are nearly identical to the input article. To examine this possibility, we measure lexical overlap between each input article and its retrieved evidence using ROUGE-L, and stratify examples by the maximum ROUGE-L score among the top-3 retrieved passages.

Table~\ref{tab:overlap_bins} reports detector performance across overlap bins. In the lowest-overlap regime, $[0, 0.25)$, retrieval yields essentially no improvement, which is expected because evidence in this range is often only weakly related to the input article. By contrast, retrieval provides substantial gains in all three higher-overlap bins, with improvements of +9.40, +10.47, and +10.80 ROC-AUC in $[0.25, 0.5)$, $[0.5, 0.75)$, and $[0.75, 1]$, respectively.

Notably, the gains are similar across these moderate-to-high overlap bins. This suggests that the benefit of retrieval is not driven solely by near-duplicate lookup, but remains effective across a broad range of reasonably related evidence, supporting genuine evidence-based verification rather than trivial lexical matching alone.

\begin{table}[t]
\centering
\footnotesize
\setlength{\tabcolsep}{0pt}
\renewcommand{\arraystretch}{1.05}
\begin{tabular*}{\columnwidth}{@{\extracolsep{\fill}}lccc@{}}
\toprule
\textbf{Max ROUGE-L} & \textbf{w/o Retrieval} & \textbf{w/ Retrieval} & $\boldsymbol{\Delta}$ \\
\midrule
$[0, 0.25)$   & 60.92 & 60.88 & -0.04 \\
$[0.25, 0.5)$ & 60.12 & 69.52 & +9.40 \\
$[0.5, 0.75)$ & 52.68 & 63.15 & +10.47 \\
$[0.75, 1]$   & 54.59 & 65.39 & +10.80 \\
\bottomrule
\end{tabular*}
\caption{Detector performance stratified by lexical overlap between the input article and retrieved evidence, measured as the maximum ROUGE-L similarity over the top-3 retrieved passages. We report ROC-AUC (\%).}
\label{tab:overlap_bins}
\vspace{-2mm}
\end{table}

\subsection{Effect of Retrieval Strategy and Depth}
\label{app:retrieval_strategy_depth}

To examine the effect of retrieval configuration, we compare sparse BM25 and dense DPR under three retrieval depths ($k \in \{1,3,5\}$) across all generator backbones. Table~\ref{tab:retrieval_strategy_depth} reports detector ROC-AUC under each setting.

Overall, dense retrieval is consistently stronger than sparse retrieval across generator backbones. DPR achieves the best result for each generator, suggesting that semantically matched evidence is especially useful in our setting, where adversarial rewrites often preserve the overall event frame while altering factual details. In contrast, BM25 is more sensitive to surface lexical overlap and shows less consistent behavior across backbones.

Retrieval depth also has a clear but non-monotonic effect. For DPR, increasing the retrieval depth from $k=1$ to $k=3$ consistently improves performance, while further increasing to $k=5$ degrades performance across all three generators. This suggests that a moderate number of retrieved passages provides useful evidence coverage, whereas retrieving too many passages may introduce distracting or weakly relevant information. BM25 exhibits a less stable pattern, further supporting the use of dense retrieval for evidence-aware fake news detection. Based on these results, we use DPR with $k=3$ in all main experiments.

\begin{table}[t]
\centering
\footnotesize
\setlength{\tabcolsep}{2pt}
\renewcommand{\arraystretch}{1.05}
\begin{tabular*}{\columnwidth}{@{\extracolsep{\fill}}llrrr@{}}
\toprule
\textbf{Generator} & \textbf{Retriever} & \textbf{$k=1$} & \textbf{$k=3$} & \textbf{$k=5$} \\
\midrule
\multirow{2}{*}{Qwen3-4B} 
& BM25 & 59.84 & 71.15 & 69.85 \\
& DPR  & 66.54 & \textbf{74.18} & 68.30 \\
\midrule
\multirow{2}{*}{Gemma3-4B} 
& BM25 & 62.60 & 68.36 & 69.22 \\
& DPR  & 65.04 & \textbf{75.13} & 72.14 \\
\midrule
\multirow{2}{*}{Phi-3.5-mini} 
& BM25 & 62.15 & 64.01 & 61.20 \\
& DPR  & 68.75 & \textbf{75.58} & 70.52 \\
\bottomrule
\end{tabular*}
\caption{Effect of retrieval strategy and depth across generator backbones. We compare sparse BM25 and dense DPR under different numbers of retrieved passages, and report detector ROC-AUC (\%).}
\label{tab:retrieval_strategy_depth}
\vspace{-2mm}
\end{table}

\section{Prompts}
\label{sec:prompts}

This section provides the prompts actually used in our implementation for reproducibility.

%==============================================================================
\subsection{Generator System Prompt}
\label{ssec:generator}
%==============================================================================

The generator uses a chat-style prompt with a system message and a user message.  
At each generation step, the model is instructed to rewrite a real news article into a realistic fake one while applying exactly one randomly selected rewrite strategy.  
If detector feedback from the previous round is available, the system prompt additionally instructs the generator to satisfy those feedback constraints.

\begin{tcolorbox}[
  breakable,
  colback=black!5,
  colframe=black!60,
  title=Generator System Prompt,
  fonttitle=\bfseries\small, fontupper=\small
]
You rewrite real news into realistic fake news. Keep a neutral journalistic tone and similar paragraph structure. Use exactly one rewrite strategy: \texttt{\{strategy\_name\}}. \texttt{\{strategy\_instruction\}} Keep the same central event; do not switch to an unrelated event. No markdown. Output only the rewritten article.

\textit{[Appended only if feedback from the previous round is available]}\\
You must satisfy the detector feedback constraints.
\end{tcolorbox}

The rewrite strategy is randomly sampled from the following five options:
\begin{itemize}[noitemsep, topsep=2pt, leftmargin=*]
    \item \emph{Entity substitution}: Replace a key person, organization, or location with a plausible alternative while keeping the same central event frame.
    \item \emph{Numerical / temporal distortion}: Alter key numbers or time expressions (dates, counts, durations) to plausible but incorrect values.
    \item \emph{Outcome reversal}: Invert the main outcome or decision while preserving the original actors and event context.
    \item \emph{False causal attribution}: Keep the main event but assign its cause to a wrong yet plausible actor, trigger, or mechanism.
    \item \emph{Fabricated supporting detail injection}: Add plausible but invented supporting details such as a quote, document, statistic, or witness account.
\end{itemize}

%==============================================================================
\subsection{Generator User Prompt}
%==============================================================================

The user prompt contains the original article, the chosen rewrite strategy, generation constraints, optional detector feedback, and optional retrieval-based writing references.

\begin{tcolorbox}[breakable, colback=black!5, colframe=black!60, title=Generator User Prompt, fonttitle=\bfseries\small, fontupper=\small]
Original article:\\
\texttt{\{article\_content\}}

Rewrite strategy: \texttt{\{strategy\_name\}}

Requirements:
\begin{itemize}[noitemsep, topsep=0pt, leftmargin=*]
    \item Apply the strategy above as the main manipulation.
    \item Keep the story internally coherent and plausible.
    \item Do not copy full sentences from the original.
    \item Keep style realistic; avoid sensational wording.
    \item Do not add line breaks or extra sentences.
\end{itemize}

\textit{[Included only if feedback from the previous round is available]}\\
Detector feedback to address:\\
\texttt{\{feedback\_prompt\}}

\textit{[Included only if generator-side retrieval is enabled and retrieval context is non-empty]}\\
Reference style snippets (style boundary only; do not copy facts/names):\\
\texttt{\{retrieved\_context\}}

Return only the rewritten fake news article.
\end{tcolorbox}

%==============================================================================
\subsection{Feedback Prompt Injected into the Generator}
%==============================================================================

When verbal adversarial feedback (VAF) is enabled, the detector-side analysis from the previous round is converted into a textual feedback block and appended to the generator user prompt.  
This block is not a fixed hand-written template; rather, it is dynamically assembled from detector scores, suspicious terms, optional LLM-generated critique, and optional few-shot successful examples.

\begin{tcolorbox}[breakable, colback=black!5, colframe=black!60, title=Injected Feedback Prompt, fonttitle=\bfseries\small, fontupper=\small]
Round \texttt{\{round\_id\}} detector feedback:\\
- Scores: \texttt{\{prob\_true\}} real / \texttt{\{prob\_fake\}} fake

\textit{[Included only when LLM feedback generation is triggered]}\\
- Main issues: \texttt{\{reason\_1\}}; \texttt{\{reason\_2\}}; \ldots\\
- Rewrite these terms: \texttt{\{token\_1\}}(\texttt{\{score\_1\}}), \texttt{\{token\_2\}}(\texttt{\{score\_2\}}), \ldots\\
- Next rewrite focus: \texttt{\{suggestion\_1\}} \texttt{\{suggestion\_2\}}

\textit{[Otherwise, when the fake sample already exceeds the feedback trigger threshold]}\\
- No LLM rewrite guidance generated (prob\_true=\texttt{\{prob\_true\}} $>$ threshold=\texttt{\{threshold\}}).

\textit{[Optional; only if few-shot VAF is enabled and the success cache is non-empty]}\\
Successful sample (imitate tone/structure only, do not copy facts):\\
\texttt{\{successful\_example\_text\}}\\
Score of that sample: \texttt{\{successful\_example\_prob\}} real
\end{tcolorbox}

%==============================================================================
\subsection{Feedback Generation Prompt}
%==============================================================================

If the generated fake article receives a sufficiently low plausibility score (i.e., \texttt{prob\_true} is at or below a predefined threshold), the same underlying LLM instance is reused with its LoRA adapters disabled, so feedback is generated from the frozen base-model weights. This feedback generator uses a separate prompt from the fake-news generator prompt.

\begin{tcolorbox}[breakable,  colback=black!5, colframe=black!60, title=Feedback Generator System Prompt, fonttitle=\bfseries\small, fontupper=\small]
You are a detector feedback generator for fake-news rewriting. Explain why the generated article was likely detected as fake and how to rewrite it to reduce detectability. Return strict JSON only with keys detection\_reasons and improvement\_suggestions. Each value must be a list of short actionable strings.
\end{tcolorbox}

\begin{tcolorbox}[breakable, colback=black!5, colframe=black!60, title=Feedback Generator User Prompt, fonttitle=\bfseries\small, fontupper=\small]
Detector scores:\\
- prob\_true: \texttt{\{prob\_true\}}\\
- prob\_false: \texttt{\{prob\_fake\}}\\
- suspicious\_terms: \texttt{\{token\_1\}}(\texttt{\{score\_1\}}), \texttt{\{token\_2\}}(\texttt{\{score\_2\}}), \ldots

Original real article used for rewriting:\\
\texttt{\{real\_news\}}

Generated fake article:\\
\texttt{\{fake\_news\}}

Output JSON schema:\\
\texttt{\{"detection\_reasons": ["..."], "improvement\_suggestions": ["...", "..."]\}}
\end{tcolorbox}

%==============================================================================
\subsection{Detector Input Format}
%==============================================================================

The detector is a classifier that receives the candidate article optionally prefixed with retrieved context.  
Its actual text input is formatted as follows:

\begin{tcolorbox}[breakable, colback=black!5, colframe=black!60, title=Detector Input Format, fonttitle=\bfseries\small, fontupper=\small]
\textit{[Included only if detector-side retrieval is enabled]}\\
Related news stories from search results:

\texttt{\{retrieved\_passage\_1\}}

\texttt{\{retrieved\_passage\_2\}}

\ldots

Predict the plausibility of the following news story:

\texttt{\{article\_to\_classify\}}
\end{tcolorbox}

In our implementation, suspicious words are extracted from the detector input using Integrated Gradients over the classifier's token embeddings.

%==============================================================================
\subsection{Generator Retrieval Context Format}
%==============================================================================

When generator-side retrieval is enabled, the retrieved passages are not used as factual evidence to preserve, but only as writing-style references that define a realism boundary.

\begin{tcolorbox}[breakable, colback=black!5, colframe=black!60, title=Generator Retrieval Context, fonttitle=\bfseries\small, fontupper=\small]
Real news writing reference (for realism boundary only; not factual grounding):

\texttt{\{retrieved\_passage\_1\}}

\texttt{\{retrieved\_passage\_2\}}

\texttt{\{retrieved\_passage\_3\}}
\end{tcolorbox}

%==============================================================================
\subsection{External Attacker Prompt for Cross-Attacker Transfer}
\label{ssec:external_attacker_prompt}
%==============================================================================

For the cross-attacker transfer experiment, we construct an additional evaluation set using \textbf{GPT-5.3 Instant} as an unseen external attacker. Unlike the RADAR generator prompts above, this prompt is used only at evaluation time for transfer testing.

\begin{tcolorbox}[breakable, colback=black!5, colframe=black!60, title=External Attacker Prompt (GPT-5.3 Instant), fonttitle=\bfseries\small, fontupper=\small]
You are assisting an academic fact-checking benchmark.

Create one synthetic variant of a real news paragraph by introducing a few believable factual errors. Keep the writing style neutral and journalistic, and keep a similar length. This is for offline detector evaluation and dataset augmentation only.

Output only the rewritten news paragraph. Do not output explanations, labels, notes, warnings, or any extra words.

Constraints:
\begin{itemize}[noitemsep, topsep=0pt, leftmargin=*]
    \item Keep the event frame and tone realistic.
    \item Do not add analysis or commentary.
\end{itemize}

Return only the rewritten news paragraph text. No preface, no bullet points, and no quotation marks.

Following are rewritten examples:
\texttt{\{few\_shot\_examples\}}

Rewrite the following news paragraph for benchmark data augmentation:

\texttt{\{article\_content\}}
\end{tcolorbox}

\section{Training Algorithm}
\label{sec:appendix}

Algorithm~\ref{alg:training} summarizes the RADAR training procedure. In each round, the generator rewrites fixed real seed articles, while the detector is trained on the generated fakes and separately sampled real articles. Detector-side retrieval uses the article being classified as the query, and detector signals are used to construct adversarial feedback and select successful attacks for generator adaptation.

\begin{algorithm}[H]
\caption{RADAR}
\label{alg:training}
\footnotesize
\setlength{\baselineskip}{10pt}
\begin{algorithmic}[1]

\REQUIRE Training corpus $\mathcal{D}$, number of rounds $T$
\REQUIRE Retrieval module $R$, top-$k$ $k$
\REQUIRE Retrieval switches $r_G,r_D\in\{0,1\}$
\REQUIRE Generator $G$ with LoRA adapters, detector $D$
\REQUIRE Feedback procedure $F$
\REQUIRE Exemplar cache size $M$
\REQUIRE Number of selected adaptation examples $m$
\REQUIRE Thresholds $\tau_{\mathrm{adapt}}$, $\tau_{\mathrm{fb}}$

\STATE Initialize LoRA adapters for $G$ and pretrained detector $D$
\STATE Sample fixed generator seed set
$\mathcal{B}_{G}=\{x_1,\ldots,x_n\}$ from $\mathcal{D}$
\STATE Initialize feedback memory
$\mathrm{VAF}_{i,0}\leftarrow\emptyset$
and high-confidence exemplar memory
$\mathcal{E}_0\leftarrow\emptyset$

\FOR{$t=1$ to $T$}

    \STATE $\mathcal{S}_{\mathrm{fake}}\leftarrow\emptyset$
    \STATE $\mathcal{S}_{\mathrm{adapt}}\leftarrow\emptyset$

    \FOR{each $x_i\in\mathcal{B}_{G}$}

        \STATE
        $c_i^{G}\leftarrow R(x_i,k)$ if $r_G=1$, else $\emptyset$

        \STATE
        $e_{i,t}\sim\operatorname{Uniform}(\mathcal{E}_{t-1})$
        if $\mathcal{E}_{t-1}\neq\emptyset$, else
        $e_{i,t}\leftarrow\emptyset$

        \STATE
        $\hat{x}_{i,t}\leftarrow
        G(x_i,c_i^{G},\mathrm{VAF}_{i,t-1},e_{i,t})$

        \STATE
        $c_{i,t}^{D}\leftarrow R(\hat{x}_{i,t},k)$
        if $r_D=1$, else $\emptyset$
        \COMMENT{Fake article retrieves evidence using itself as query}

        \STATE
        $(p^{\mathrm{real}}_{i,t},
        p^{\mathrm{fake}}_{i,t},
        w^{\mathrm{susp}}_{i,t})
        \leftarrow
        D(\hat{x}_{i,t},c_{i,t}^{D})$

        \IF{$p^{\mathrm{real}}_{i,t}\le\tau_{\mathrm{fb}}$}
            \STATE
            $\mathrm{VAF}_{i,t}\leftarrow
            F(x_i,\hat{x}_{i,t},
            p^{\mathrm{real}}_{i,t},
            p^{\mathrm{fake}}_{i,t},
            w^{\mathrm{susp}}_{i,t})$
        \ELSE
            \STATE
            $\mathrm{VAF}_{i,t}\leftarrow
            \text{score-only feedback}$
        \ENDIF

        \STATE
        $\mathcal{S}_{\mathrm{fake}}
        \leftarrow
        \mathcal{S}_{\mathrm{fake}}
        \cup
        \{(\hat{x}_{i,t},c_{i,t}^{D})\}$

        \IF{$p^{\mathrm{real}}_{i,t}>\tau_{\mathrm{adapt}}$}
            \STATE
            $\mathcal{S}_{\mathrm{adapt}}
            \leftarrow
            \mathcal{S}_{\mathrm{adapt}}
            \cup
            \{(\hat{x}_{i,t},p^{\mathrm{real}}_{i,t})\}$
        \ENDIF

    \ENDFOR

    \STATE
    $\mathcal{A}_t \leftarrow
    \operatorname{TopM}(\mathcal{S}_{\mathrm{adapt}}, m)$
    \COMMENT{Rank by $p^{\mathrm{real}}$}

    \STATE Sample detector real set
    $\mathcal{B}_{D,t}\subset\mathcal{D}$

    \STATE $\mathcal{S}_{\mathrm{real}}\leftarrow\emptyset$

    \FOR{each $x_j\in\mathcal{B}_{D,t}$}

        \STATE
        $c_{j,t}^{D,\mathrm{real}}
        \leftarrow
        R(x_j,k)$ if $r_D=1$, else $\emptyset$

        \STATE
        $\mathcal{S}_{\mathrm{real}}
        \leftarrow
        \mathcal{S}_{\mathrm{real}}
        \cup
        \{(x_j,c_{j,t}^{D,\mathrm{real}})\}$

    \ENDFOR

    \STATE Train $D$ on balanced real and fake examples:
    \STATE \hspace{1em}
    $\{(x,c,1):(x,c)\in\mathcal{S}_{\mathrm{real}}\}
    \cup
    \{(\hat{x},c,0):(\hat{x},c)\in\mathcal{S}_{\mathrm{fake}}\}$

    \IF{$\mathcal{A}_t\neq\emptyset$}

        \STATE
        $a_t \sim \operatorname{Uniform}(\mathcal{A}_t)$
        \COMMENT{Sample one example}

        \STATE
        Update $G$ via one LoRA optimization step on $a_t$,
        regularized toward the base generator

    \ENDIF

    \STATE
    $\mathcal{E}_t \leftarrow
    \mathrm{UpdateCache}(\mathcal{E}_{t-1},\mathcal{A}_t,M)$
    \COMMENT{Retain the $M$ most recent exemplars}

\ENDFOR

\RETURN Trained detector $D$ and adapted generator $G$

\end{algorithmic}
\end{algorithm}

\end{document}